\documentclass[10pt,twocolumn,letterpaper]{article}

\usepackage{cvpr}
\usepackage{times}
\usepackage{epsfig}
\usepackage{graphicx}
\usepackage{amsmath}
\usepackage{amssymb}

\usepackage{cvpr}
\usepackage{times}
\usepackage{epsfig}
\usepackage{graphicx}
\usepackage{amsmath}
\usepackage{amssymb}
\usepackage{url}
\usepackage{times}  
\usepackage{helvet}  
\usepackage{courier}  
\usepackage{url}  
\usepackage{graphicx}  
\usepackage[utf8]{inputenc} 
\usepackage[T1]{fontenc}    
\usepackage{url}            
\usepackage{booktabs}       
\usepackage{amsfonts}       
\usepackage{nicefrac}       
\usepackage{microtype}      

\usepackage{wrapfig}
\usepackage{picinpar}
\usepackage{cutwin}

\usepackage{times}
\usepackage{epsfig}
\usepackage{graphicx}
\usepackage{amsmath}
\usepackage{amssymb}

\usepackage{caption}
\usepackage{enumitem}
\usepackage{multirow}
\usepackage{booktabs}
\usepackage{subfigure}

\usepackage{amsfonts,amssymb}
\usepackage{graphicx}
\usepackage{algorithm}
\usepackage{algorithmic}

\usepackage{color}

\usepackage[breaklinks=true,bookmarks=false]{hyperref}
\cvprfinalcopy 

\def\cvprPaperID{242} 

\begin{document}

\title{Adaptively Connected Neural Networks}

\author{Guangrun Wang\\
Sun Yat-sen University\\
Guangzhou\\
{\tt\small wanggrun@mail2.sysu.edu.cn}
\and
Keze Wang\\
University of California, Los Angeles\\
Los Angeles\\
{\tt\small kezewang@gmail.com}
\and
Liang Lin\\
Sun Yat-sen University\\
Guangzhou\\
{\tt\small linliang@ieee.org}
}

\maketitle


\begin{abstract}
This paper presents a novel adaptively connected neural network (ACNet) to improve the traditional convolutional neural networks (CNNs) {in} two aspects. First, ACNet employs a flexible way to switch global and local inference in processing the internal feature representations by adaptively determining the connection status among the feature nodes (e.g., pixels of the feature maps) \footnote{In a computer vision domain, a node refers to a pixel of a feature map{, while} in {the} graph domain, a node denotes a graph node.}. We can show that existing CNNs, the classical multilayer perceptron (MLP), and the recently proposed non-local network (NLN) \cite{nonlocalnn17} are all special cases of ACNet. Second, ACNet is also capable of handling non-Euclidean data. Extensive experimental analyses on {a variety of benchmarks (i.e.,} ImageNet-1k classification, COCO 2017 detection and segmentation, CUHK03 person re-identification, CIFAR analysis, and Cora document categorization) demonstrate that {ACNet} cannot only achieve state-of-the-art performance but also overcome the limitation of the conventional MLP and CNN \footnote{Corresponding author: Liang Lin (linliang@ieee.org)}. The code is available at \url{https://github.com/wanggrun/Adaptively-Connected-Neural-Networks}.
\end{abstract}

\section{Introduction}
\label{sect:intro}

Artificial neural networks have been extensively studied and applied over the past three decades, achieving remarkable accomplishments in artificial intelligence and computer vision. Among such networks, two types of neural networks have had a large impact on the research community. The first type is the multi-layer perceptron (MLP), which first became popular and effective via the development of the back-propagation training algorithm~\cite{BP,BP2}. However, since each neuron of the hidden layer in MLP is assigned with a private weight,  the network parameters of MLP usually have a huge number and can be easily overfitted during the training phase. Moreover, MLP has difficulty in representing the spatial structure of 2D data (e.g., images). The second type is convolutional neural networks (CNNs)~\cite{lecun1990handwritten}. Motivated by the biological visual cortex model, CNNs propose to group adjacent neurons to share identical weights and represent 2D data by capturing the local pattern (i.e., receptive field) of each neuron.

\begin{figure}[t]
\centering
\includegraphics[width = 1.1\columnwidth]{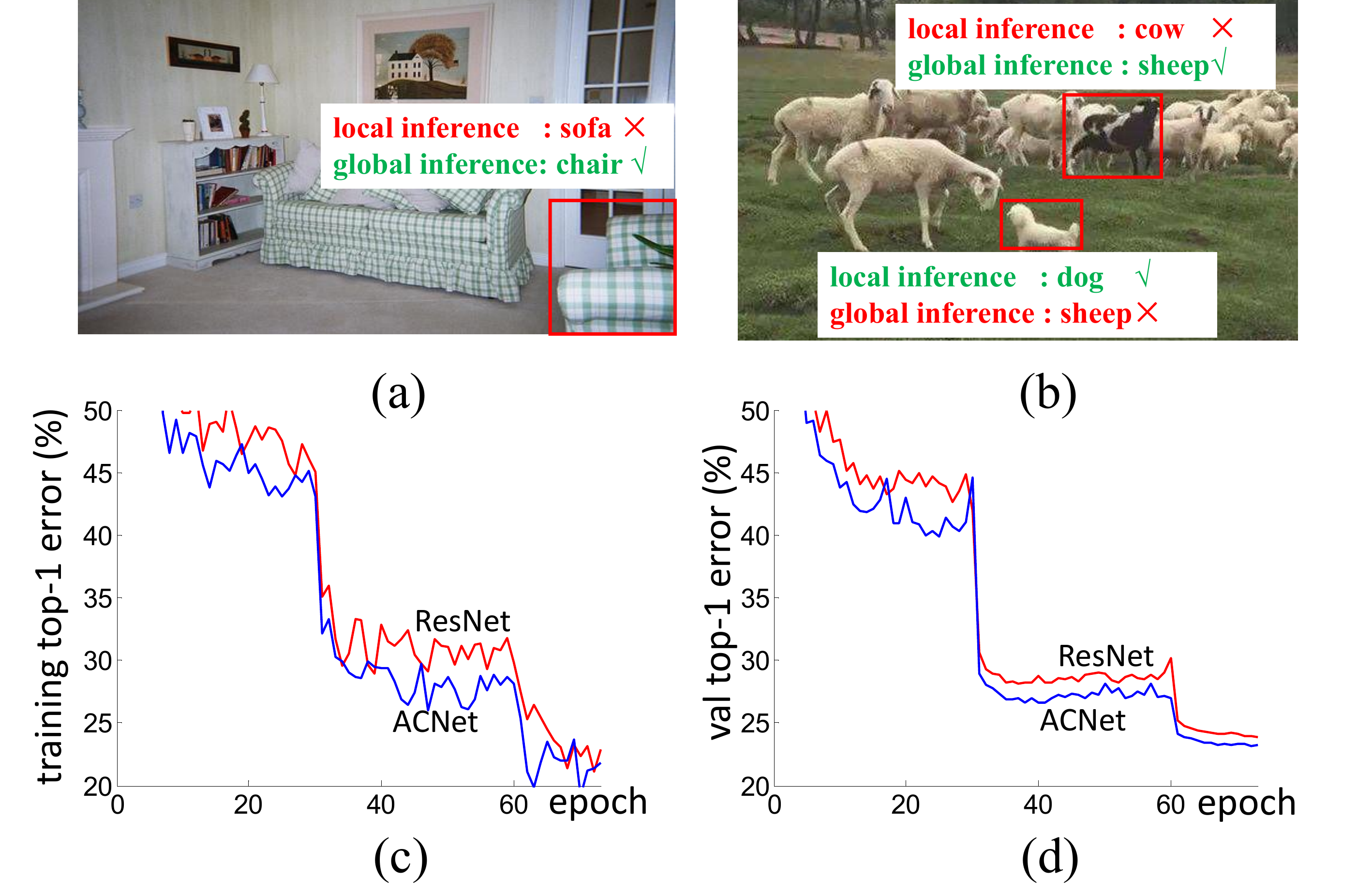}
\caption{\small{Some pixels prefer global dependencies, while others prefer local inference. For example, without global inference we cannot recognize the \emph{chair} in (a). While in (b), the representation capacity of the \emph{dog} is weakened by global information. Thanks to the adaptively determining the global/local inference, our ACNet achieves lower top-1 training/validation error than ResNet on ImageNet-$1k$ shown in (c) and (d).}}\label{fig:intro}
\vspace{-11pt}
\end{figure}

Although CNNs have been proven to be significantly superior over MLP, they have two drawbacks, as highlighted by~\cite{capsule}.
On one hand, due to only abstracting information from local neighborhood pixels, the convolution operation inside each layer of CNNs does not have the ability of global inference.
Consequently, convolution operations have difficulties in recognizing objects with similar appearances.
For example, a convolution operation cannot distinguish the difference between the chair and the sofa in Fig.\ref{fig:intro} (a) which share the same appearance.
In practice, CNN captures the global dependencies by stacking a number of local convolution operations, which still have several limitations, such as computational inefficiency, optimization difficulty, and message passing inefficiency \cite{nonlocalnn17}.
On the other hand, unlike MLP, conventional CNNs cannot be directly applied for non-Euclidean data (e.g., graph data), which are quite common in the area of machine learning.

\begin{figure}[t]
\centering
\includegraphics[width = 1\columnwidth]{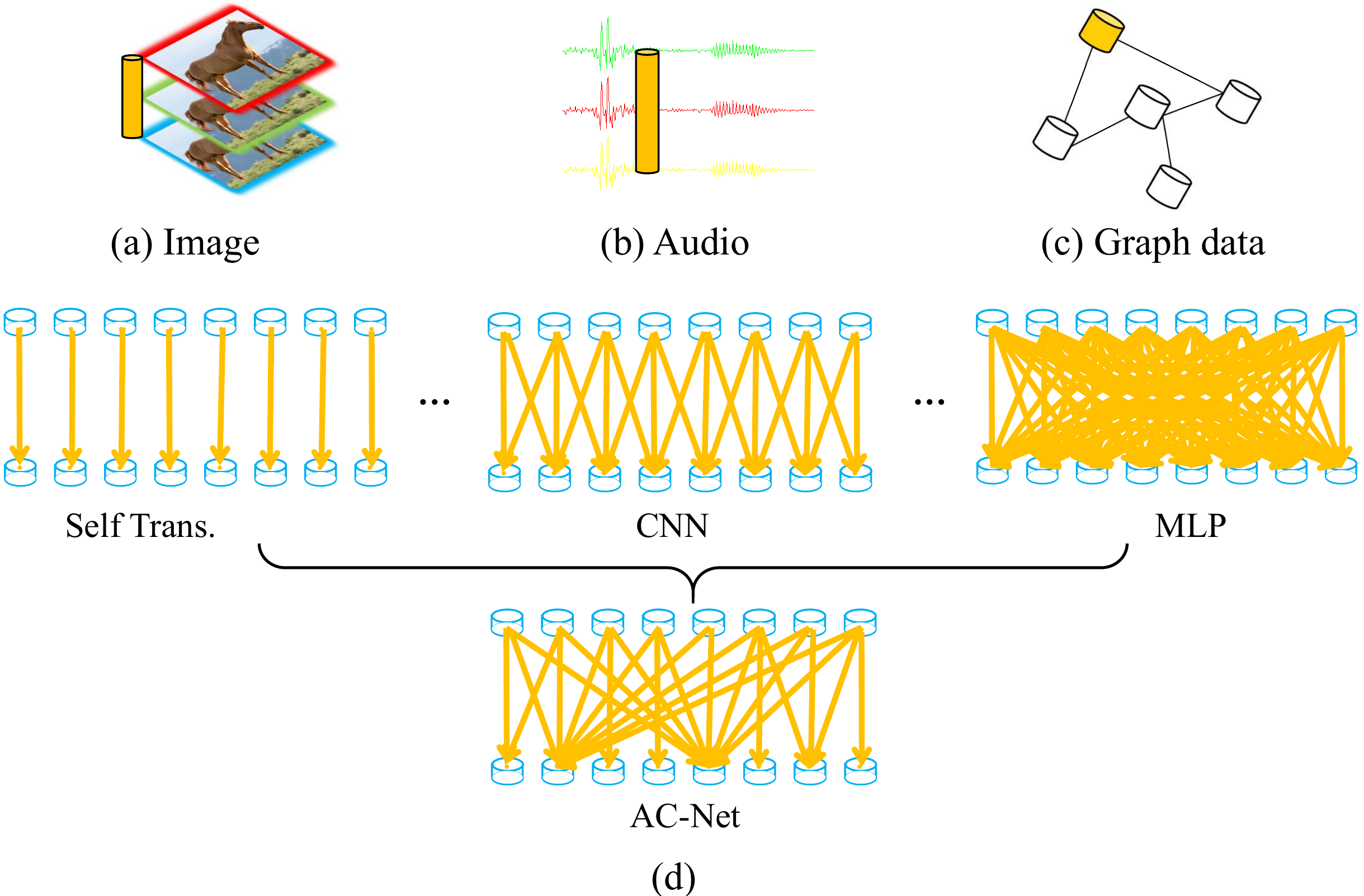}
\caption{\small{``Nodes'' are presented in form of orange cylinder in (a) an image, (b) an audio, and (c) a general graph. (d) ACNet can be considered as a generalization of MLP and CNN on these ``nodes''.}}\label{fig:node}
\vspace{-11pt}
\end{figure}

To tackle the locality problem in CNNs,
the recently proposed non-local network \cite{nonlocalnn17} (denoted as fully non-local networks) imposes global dependencies to all the feature nodes.
However, empirically we observe degradations in fully non-local networks: as the non-locality of the network increases, both the training and validation accuracies degrade for {the} ImageNet-$1k$ classification.
We conjecture the degradation due to over-globalization. Specifically, the \emph{dog} in Fig.\ref{fig:intro} (b) is easy to recognize if we only perform the local inference, while it can be misclassified as a cow when only performing the global inference.
Intuitively, although quite challenging, it is necessary to jointly consider the global and local inference from image-aware (Fig.\ref{fig:intro}(a)) or even node-aware (pixel-aware) (Fig.\ref{fig:intro}(b)) perspective.

There have been many other recent attempts to address the aforementioned issues raised by CNNs and have achieved promising results \cite{capsule,matrixCapsule}.
However, all of these methods are either over-localized or over-globalized.
In contrast, this work focuses on developing a simple and general adaptively-connected neural network (ACNet) to adaptively capture the global and local dependencies,
which inherits the strengths of both MLP and CNNs and overcomes their drawbacks. Thanks to the adaptively determining the global/local inference, our ACNet achieves lower top-1 training/validation error than ResNet on ImageNet-$1k$ (see Fig. \ref{fig:intro}(c) and (d)).

ACNet first defines a simple yet basic unit named ``node'', which is a unit of vectors in meta-data.
As depicted in Fig.\ref{fig:node}, a node may be seen as a pixel of an image (Fig. \ref{fig:node}(a)), a sampling of an audio (Fig. \ref{fig:node}(b)), and a node of a general graph (Fig. \ref{fig:node}(c)).
Given the input data, ACNet adaptively is trained to search the optimal connection for each node, i.e., the connection $\subseteq$ connecting $\{$the node itself, its neighbor nodes, all possible nodes$\}$. Keep in mind that different nodes are connected adaptively, i.e., some nodes may be conjectured to themselves, some nodes may relate to its neighborhood, while other nodes have the global vision. Therefore, our ACNet can be considered as a generalization of CNN and MLP (Fig. \ref{fig:node} (d)).
Note that, searching the optimal connections is differential by learning the importance degrees for different kinds of connections, which can be optimized via back-propagation.

The \textbf{main} contributions of this paper are summarized as follows. Firstly, we propose a conceptually general yet powerful network, which learns to switch global and local inference for general data (i.e. both Euclidean and non-Euclidean data) in a flexible parameter saving manner. Secondly, to the best of our knowledge, our proposed ACNet is the first one who is capable of inheriting the strength of both MLP and CNN while overcoming their drawbacks on a variety of computer vision and machine learning tasks, i.e., image classification on ImageNet-1k/CIFARs, object detection and segmentation on COCO 2017, person re-identification on CUHK03, and document categorization on Cora.

\section{Related Work}

Although significant progress has been achieved in the architecture design of CNNs from LeNet~\cite{lenet} to more recent deep and powerful networks (e.g., ResNet~\cite{resnet}), evolving the structure of CNNs to overcome their drawbacks is also quite crucial and a long-standing problem in machine learning (\eg \cite{li2017aognets}). This issue motivates many researchers to extend CNNs to obtain different receptive fields \cite{deformableCNN17ICCV}. Specifically, Dai {\em et al.}~\cite{deformableCNN17ICCV} proposed to enhance the transformation modeling capability of CNNs by introducing learnable offsets to augment the spatial sampling locations within the feature map. Chen {\em et al.}~\cite{chen2017rethinking} revisited atrous convolution, a powerful tool to explicitly adjust filter's field-of-view as well as control the resolution of feature responses computed by DNNs, in the application of semantic image segmentation. Peng {\em et al.} \cite{peng2017large} proposed to use the large kernel filter and effective receptive field for semantic segmentation. Sabour {\em et al.}~\cite{capsule} proposed employing a group of neurons named a capsule to represent the instantiation parameters of a specific type of entity, such as an object and an object part. 
Building upon the work of \cite{capsule}, Hinton {\em et al.}~\cite{matrixCapsule} further presented a new type of capsule that has a logistic unit to represent the presence of an entity and a 4$\times$4 pose matrix to represent the pose of that entity. Motivated by the self-attention mechanism~\cite{selfattention17NIPS}, Wang {\em et al.}~\cite{nonlocalnn17} incorporated non-local operations into CNNs as a generic family of building blocks for capturing long-range dependencies. Similarly, PSANet \cite{zhao2018psanet} is built upon NLN by introducing a position encoding to each pixel; GloRe \cite{chen2018graph} improves on NLN in a way of using a graph-CNN to capture the global dependencies. Although these methods achieved promising results, their performances are limited due to the over-localization or over-globalization of the internal feature representation.

Moreover, several limited attempts~\cite{kipf2017semi,graphcnn,gao2018large,zhuang2018dual} have been made to extend CNNs for handling graph data. For instance, Kipf {\em et al.}~\cite{kipf2017semi} presented a layer-wise propagation rule for CNNs to operate directly on graph-structured data. Such {\em et al.}~\cite{graphcnn} defined filters as polynomials of functions of the graph adjacency matrix for unstructured graph data. However, these variants of CNNs pay close attention to bridge the gap between the graph structure of network inputs and the general graph data. The global inference inside the internal representation are ignored. 

Our work is also related to the fully-connected neural networks (i.e. multilayer perceptron, or MLP), the densely connected neural networks \cite{huang2017densely}, and the skip-connection neural networks (e.g. UNet \cite{ronneberger2015u}, ResNet \cite{he2016deep}), sharing the goal of finding an effective connection for the neural networks. However, the connections in our ACNet are automatically learned and adaptative to the data, while the connections in existing methods are fixed and handcrafted.

\section{Adaptive-Connected Neural Networks}\label{sect:method}


In this section, we first present the formulation of our proposed ACNet. Then, we discuss the relations between our ACNet and three most representative prior works, i.e., MLP, CNN, and NLN.  Actually, they are special cases of our ANN. Moreover, we have also generalized our ANN for non-Euclidean data. Finally, we present the details of training, testing, and implementing our ACNet.


\subsection{Formulation}
\label{sec:formulation}
Suppose $x$ denotes the input signal (e.g., images, voices, graph matrices or their features). We propose to obtain the corresponding output signal as follows:\begin{small}\begin{equation}\label{eq:ann}
\begin{aligned}
\mathbf{y}_{i} & = \alpha_i \sum_{j=i} \mathbf{x}_j \mathbf{u}_{ij} + \beta_i \sum_{j\subseteq N(i)} \mathbf{x}_j \mathbf{v}_{ij} + \gamma_i \sum_{\forall j} \mathbf{x}_j \mathbf{w}_{ij},    
\end{aligned}
\end{equation}\end{small}where $y_i$ implies the $i$-th output node (e.g., the $i$-th pixel of a feature map) of the output signal, and $j$ is the index of some possible nodes related to the $i$-th node. Actually, the $j$-th node belongs to three different sets, including \{the $i$-th node itself\}, \{the neighborhood $N(i)$ of the $i$-th node\}, and \{all possible nodes\}.
These three sets indicate three different modes of inference: self transformation, local inference, and global inference, respectively. Moreover,
$\mathbf{u}_{ij}$, $\mathbf{v}_{ij}$ and $\mathbf{w}_{ij}$ refer to the learnable weights between the $i$-th and $j$-th nodes for the three different sets, respectively. Note that the biases are omitted for notation simplification.

{ACNet} switches among different inference modes by adaptively learning $\alpha$, $\beta$ and $\gamma$ in Eqn.\ref{eq:ann}, which are importance degrees used to weighted average the modes. Note that, $\alpha$, $\beta$ and $\gamma$ can be simple scalar variables, which are shared across all channels. We force $\alpha + \beta + \gamma = 1$, and $\alpha, \beta, \gamma \in [0,1]$, and define\begin{small}\begin{equation}\label{eq:switch}
\alpha = \frac{e^{\lambda_{\alpha}}}{e^{\lambda_{\alpha}} + e^{\lambda_{\beta
}} + e^{\lambda_{\gamma}}}.
\end{equation}\end{small}Here $\alpha$ is computed by using a softmax function with $\lambda_{\alpha}$ as the control parameter, which can be learned by the standard back-propagation (BP). Similarly, $\beta$ and $\gamma$ are defined by using another parameters $\lambda_{\beta}$ and $\lambda_{\gamma}$, respectively. Note that the third term $\sum_{\forall j} \mathbf{x}_j \mathbf{w}_{ij}$ in Eqn.\ref{eq:ann} is quite {computational consuming}, because it equals to a fully-connected layer with large feature maps as input, leading to potential overfitting. To overcome this shortcoming, the $\mathbf{x}$ is first transformed by an average pooling operation for downsampling in practice before being fed to calculate $\sum_{\forall j} \mathbf{x}_j \mathbf{w}_{ij}$. Finally, the obtained $\mathbf{y}$ in Eqn.\ref{eq:ann} {can} be activated by a non-linear function $f(\cdot)$, such as BatchNorm+ReLU.

Actually, if $\alpha,\beta,\gamma$ are formulated as scalar variables, the connection for adaptively determining the global/local inference is an average connection over the whole dataset. To enable node-aware connection for each node (e.g., a pixel), $\alpha,\beta,\gamma$ can be also formulated as sample-dependent ones:\begin{small}\begin{equation}\label{eq:unique}
\gamma_i = \gamma_i(\mathbf{x}) = \mathbf{w}_{\gamma_i,2}f(\mathbf{w}_{\gamma_i,1}\bigg[\sum_{j=i} \mathbf{x}_j \mathbf{u}_{ij};\sum_{j\subseteq N(i)} \mathbf{x}_j \mathbf{v}_{ij};\sum_{\forall j} \mathbf{x}_j \mathbf{w}_{ij}\bigg]),
\end{equation}\end{small}where $[;;]$ denotes a concatenation operation and $\mathbf{w}_{\gamma, \cdot}$ denotes a linear transformation. $\alpha$ and $\beta$ are defined in the similar way, which are omitted here. In the experimental section we will show that the above two kinds of formulation have the similar performance.

\subsection{Relation to Prior Works}
\label{sec:relation}

\textbf{CNN.} We take CNN as an illustrative example. For notation simplification, we omit the non-linear activation $f$, which does not affect the derivation process of the formulation. Let $\mathbf{x}$ be the input data {represented} by a 3D tensor ($C, H, W$). Let $\mathbf{x}_{i}$ and $\mathbf{y}_{i}$ be a node (pixel) of the input data and the output respectively, where $i,j\in [1, H\times W]$. Then a general 3$\times$3 convolution can be formulated as\begin{small}\begin{equation}\label{eq:cnn}
\begin{aligned}
\mathbf{y}_{i} & =  \sum_{j\subseteq S} \mathbf{x}_j \mathbf{v}_{ij} \\
\end{aligned}
\end{equation}\end{small}where $S$ is the set {that} containing the nodes which have interactions with the given $i$-th node. Specifically, $S$ denotes the set of eight neighbors for the $i$-th node, in addition to the $i$-th node itself, i.e., $S = \{i - W - 1, i - W, i - W + 1, i - 1, i, i + 1, i + W - 1, i + W, i + W + 1\}$.

\textbf{MLP.} MLP shares the formulation of Eqn.\ref{eq:cnn}, but it uses different sets of nodes to perform the linear combination. In other words, MLP enables more nodes to interact with the given $i$-th node, performing a global inference. For MLP, $S = \{1, 2, 3, \ldots, H\times W\}$.

In summary, ACNet can be seen as a pure data-driven combination of CNN and MLP, fully exploiting the advantage of these two kinds of basic neural networks. For instance, let $\alpha = 0, \beta = 1, \gamma = 0$ in Eqn.\ref{eq:ann}, ACNet degrades into CNN; let $\alpha = 0, \beta = 0, \gamma = 1$ in Eqn.\ref{eq:ann}, ACNet degrades into MLP.
More importantly, ACNet dynamically switches between them by learning $\alpha$, $\beta$ and $\gamma$, providing more reasonable inferences. This allows us to build a richer hierarchy that combines both global and local information adaptively.

\textbf{NLN.} NLN also shares the formulation of Eqn.\ref{eq:cnn}, with $S = \{1, 2, 3, \ldots, H\times W\}$, which is similar to MLP. However, there is a limitation in NLN. The $\mathbf{v}_{ij}$ in NLN is obtained by computing the similarity of the $i$-th and the $j$-th nodes, which is very computation-consuming and easy to overfit. Therefore, NLN is rarely employed {for} image classification tasks. {Instead of directly computing $v_{ij}$, our proposed ACNet 
absorbs the advantage of MLP (i.e.,  regarding $v_{ij}$ as a learnable weight) and tackles its heavy computation problem by employing downsampling operation to perform the global inference.} The relations between ACNet and prior works have been summarized in Fig. \ref{fig:node} (d).


\subsection{Generalization to Non-Euclidean Data}
\label{sec:generatlization}
We present the difference between Euclidean and non-Euclidean data, and then give a general definition of ACNet to handle both Euclidean and non-Euclidean data.

Euclidean data include the image, audio, and video, while non-Euclidean data contains graph and manifold.
The difference is that Euclidean data are structured and non-Euclidean data are unstructured. Mathematically, for Euclidean data, we can denote the neighborhood of the $i$-th node in Eqn. \ref{eq:ann} as $N(i) = \{i - W - 1, \ldots, i + W + 1\}$, representing the \{upper left, ..., low right \} neighbors. But for non-Euclidean data we have difficulties. Besides, each node in Euclidean data has a fixed number of neighbors, while the number of neighbors is flexibly adapted to non-Euclidean data. Consequently,
there is a gap in using Eqn.\ref{eq:ann} between Euclidean and non-Euclidean data.
For Euclidean data $\mathbf{v}_{ij}$ {has} different {values} at different $j$. But for non-Euclidean data $\mathbf{v}_{ij}$ is shared among different $j$ in Eqn.\ref{eq:ann}. This weakens the representation capacity for non-Euclidean data due to the lack of position encoding. The similar phenomenon also occurs in $\mathbf{w}_{ij}$.

To fill the gap, Eqn. \ref{eq:ann} is rewritten into a general form:\begin{small}\begin{equation}\label{eq:gann}
\begin{aligned}
\mathbf{y}_{i} & = \alpha_i \sum_{j=i} \mathbf{x}_j \mathbf{u} + \beta_i \sum_{j\subseteq N(i)} p_{ij}(\mathbf{x}_j \mathbf{v}) + \gamma_i \sum_{\forall j} q_{ij}(\mathbf{x}_j \mathbf{w}).
\end{aligned}
\end{equation}\end{small}where $\mathbf{u}$, $\mathbf{v}$, and $\mathbf{w}$ are shared among all kinds of $j$, which may be considered as $1\times1$ convolution in computer vision. Note that here $\alpha,\beta,\gamma$ is defined by using Eqn. \ref{eq:unique}. In compensation for the information loss in local structure, another two position encoding functions, i.e., $p_{ij}$ and $q_{ij}$, are proposed to encode the index. These functions are just simple linear transformations using constant Gaussian noise. Specifically,\begin{small}\begin{equation}\label{eq:position}
p_{ij}(\mathbf{x}_j \mathbf{v})  =  \mathbf{x}_j \mathbf{v} \zeta_{ij}, ~~~~q_{ij}(\mathbf{x}_j \mathbf{w})  =  \mathbf{x}_j \mathbf{w} \xi_{ij}
\end{equation}\end{small}where $\zeta_{ij}$ and $\xi_{ij}$ are constant variables sampled from a Gaussian noise.

\textbf{Remark 1.} \emph{Let $\zeta_{ij}$ and $\xi_{ij}$ in Eqn.\ref{eq:position} be learnable parameters instead of constant variables, then Eqn.\ref{eq:position} turns out to be Eqn.\ref{eq:ann}.}

Remark 1 reveals that Eqn.\ref{eq:gann} is a lightweight version of Eqn.\ref{eq:ann}, because a number of parameters are represented as constant variables in Eqn.\ref{eq:gann}, exception the $1\times1$ convolution kernels $\mathbf{u}$, $\mathbf{v}$, and $\mathbf{w}$.
In the experimental section we will show that compared to the state-of-the-art CNNs that use large kernels, ACNet with considerably fewer parameters can also achieve their strengths in feature learning, by only exploiting highly efficient $1\times1$ convolution operations.

\subsection{Training, Inference, and Implementation}

\textbf{Training \& Inference.} Let $\Theta$ be a set of network parameters (e.g. convolution filters and fully-connected weights) and $\Phi$ be a set of control parameters that control the network architecture. In ACNet, we have $\Phi = \{ \lambda_{\alpha}, \lambda_{\beta} , \lambda_{\gamma} \}$. Training an ACNet network is to minimize a loss function $\mathcal{L}(\Theta, \Phi)$, where $\Theta$ and $\Phi$ can be optimized jointly by back-propagation (BP).
ACNet is tested in the same way as standard networks such as CNN and MLP. 


\textbf{Compatibility with CNN Tricks and Techniques.} Our proposed ACNet is quite compatible with most existing tricks and techniques for CNNs. For instance, through embedding a batch normalization~\cite{ioffe2015batch} layer into every non-linear mapping function $f(\cdot)$, our ACNet can support a large learning rate for high learning efficiency. Meanwhile, we can also exploit the residual connection strategy~\cite{resnet} to create a short-cut connection for each layer inside our ACNet.

\textbf{Implementation.} ACNet can be easily implemented by using the existing software such as TensorFlow and PyTorch. The backward computation of ACNet can be obtained by automatic differentiation techniques (AD) in these software. Without AD, ACNet can also be implemented by regarding $\Phi = \{ \lambda_{\alpha}, \lambda_{\beta} , \lambda_{\gamma} \}$ as learnable parameters.

\section{Experiments}
This section presents the main results of ACNet in multiple challenging problems and benchmarks, such as ImageNet-$1k$ classification \cite{russakovsky2015imagenet}, COCO 2017 detection and segmentation \cite{lin2014microsoft}, CUHK03 person re-identification \cite{li2014deepreid}, CIFAR \cite{krizhevsky2009learning} classification, and Cora document categorization \cite{sen2008collective}, where the effectiveness of ACNet is demonstrated by comparing with {the} existing state-of-the-art CNNs/NLNs.

\subsection{ImageNet-$1k$ Classification}
We first compare our ACNet with the most representative CNNs/NLNs
on the ImageNet classification dataset of 1k categories. All the models are trained on the 1.28M training images and evaluated on the 50k validation images. Our baseline model is the representative ResNet50. We examine top-1 accuracy on the 224$\times$224 single/center-crop-single-scale images. Note that the top-1 accuracies of the baseline approximately equals to the official results and the model zoo \footnote{\url{https://github.com/Cadene/pretrained-models.pytorch}} (Caffe; Tensorflow; Pytorch). CNN-ResNet50 is exactly the original ResNet50. For ACNet-ResNet50, all the $3\times3$ convolution in CNN-ResNet50 are replaced with ACNet layers. And for NLN-ResNet50, the non-local operations are attached to every $3\times3$ convolution in CNN-ResNet50.

\begin{table}
 \caption{\small{Comparison of ImageNet \emph{val} top-1 accuracies and parameter numbers on ResNet50. \textbf{ACNet$\ddag$:} pixel-aware ACNet using Eqn. \ref{eq:unique}; \textbf{ACNet:} dataset-aware ACNet with $\alpha,\beta,\gamma$ being scalar variables; }}
\small
  \centering
   \begin{tabular}{l|cc}
    \toprule
      & top-1 accuracies & \#params\\
    \midrule
    CNN-ResNet50  & 76.4$^{\uparrow0.0}$ & 25.56M$^{\times1.00}$ \\
    \textbf{ACNet-ResNet50}   & \textbf{77.5$^{\uparrow1.1}$} & 29.38M$^{\times1.15}$ \\
    \textbf{ACNet$\ddag$-ResNet50}   & \textbf{77.5$^{\uparrow1.1}$} & 31.85M$^{\times1.25}$ \\
    \textbf{generalized ACNet}  & \textbf{76.2$^{\downarrow0.2}$} & \textbf{19.80M$^{\times0.77}$}\\
    \bottomrule
  \end{tabular}
  \label{tab:baseline}
\end{table}

\begin{table}
\caption{\small{{Comparison between ACNet-Resnet50 and CNN-ResNet60  in terms of ImageNet val top-1 accuracies and parameter numbers.}}}
\small
\centering
\begin{tabular}{l|cc}
\toprule
 & top-1 accuracies (\%) & \#params\\
\midrule
CNN-ResNet60 & 76.7$^{\uparrow0.0}$ &  30.03M$^{\times1.00}$\\
\textbf{ACNet-ResNet50}  & \textbf{77.5}$^{\uparrow0.8}$ & \textbf{29.38M}$^{\times0.98}$\\
\bottomrule
\end{tabular}\label{tab:resnet60}
\vspace{-11pt}
\end{table}

\textbf{Classification accuracies.} 
The comparison results of top-1 validation accuracies are illustrated in Table 1. As depicted, our ACNet-ResNet50 performs approximately 1.1\% better than the compared CNN-ResNet50 (77.5\% vs 76.4\%). The training and validation curves in Fig. \ref{fig:intro} (c) and (d) also show the sustainable competitive advantage of our ACNet-ResNet50 over CNN-ResNet50. This improvement is quite significant due to the challenge of ImageNet-$1k$. 

%

The superior performance of our ACNet is attributed to two reasons.
First, ACNet adaptively performs global and local inference for different pixels of internal feature maps from each layer, leading to a flexible discriminative representation learning fashion, which contributes to capturing the local and global dependencies for improving classification accuracy.
Second, the mechanism of ACNet may implicitly act as comprehensive data-driven ensembling, which aggregates the advantage of both global and local information.

\textbf{Pixel-aware Connection.} As is mentioned in Section \ref{sec:formulation}, different pixels can have different pixel-aware connection by using Eqn. \ref{eq:unique}. We report the accuracies of pixel-aware and dataset-aware connection in Table \ref{tab:baseline} respectively. For the pixel-aware connection, we let $\alpha=0, \beta = 1$ and only learn $\gamma$ to save parameters and memory. The results show that these two kinds of connection have the same top-1 accuracy. While the pixel-aware connection has more parameters (31.85M vs 29.38M).

\textbf{Extra Parameters.} In fact, ACNet has introduced extra parameters by 0.15$\times$ (29.4M \emph{vs} 25.6M, Table \ref{tab:baseline}). The extra parameters are from the global inference ( i.e. $\sum_{\forall j} \mathbf{x}_j \mathbf{w}_{ij}$ in Eqn.\ref{eq:ann}). Thanks to downsampling operation, the extra parameters only introduce negligible computation time and memory usage, which will be examined later. To eliminate the confounding factor of extra parameters and justify the gain of ACNet, we present more comparisons:
\begin{enumerate}
\item We compare ACNet-ResNet50 with CNN-ResNet60, which has the same level of parameters. The result in Table \ref{tab:resnet60} shows that ACNet-ResNet50 obtains a slightly higher accuracy (77.5\% vs 76.7\%) than CNN-ResNet60, demonstrating the superiority of ACNet over CNN with the nearly same number of parameters.
\item The general form of ACNet is also compared with CNN. As is discussed in Sect.~\ref{sec:generatlization}, ACNet can be rewritten to a general form for supporting both Euclidean and non-Euclidean data. Remark \textcolor{red}{1} in Sect.~\ref{sec:generatlization} reveals that the general form is much more ($0.77\times$) lightweight. The experimental result in Table \ref{tab:resnet60} confirms this remark, and further shows that ACNet with considerably fewer parameters can also achieve their strengths in feature learning(76.2\%vs76.4\%), by only exploiting highly efficient $1\times1$ convolution operations.
\end{enumerate}

\begin{table}
\caption{\small{Computational complexity analysis on the ImageNet-$1k$.}}
\scriptsize
\centering
\begin{tabular}{l|cccc}
\toprule
{Networks} &  CNN-ResNet50 & NLN-ResNet50 & \textbf{ACNet-ResNet50} & \\
\midrule
Speed$^{\text{images/sec}}$ & 198$^{\times\text{1.00}}$ & nan & \textbf{144}$^{\times\text{0.77}}$ \\
Memory$^{\text{GB}}$ & 8.579$^{\times1}$ & out of memory & \textbf{8.580}$^{\times1}$ \\
\bottomrule
\end{tabular}\label{tab:cmp_cost}
\vspace{-11pt}
\end{table}
\noindent

\textbf{Computation Complexity.} Table \ref{tab:cmp_cost} reports the computation complexity of ACNet, CNN and NLN. For a fair comparison, all
methods are trained in the same desktop with 8 Titan Xp GPUs.
We observe that ACNet and CNN have similar computational costs. 
Specifically, the memory consumption of both ACNet and CNN are the same, i.e. 8.6GB.
But the speed of ACNet is slightly slower than CNN (144 vs 198 images/second/GPU).
As a comparison, NLN is intractable because NLN requires a vast amount of memory to calculate the similarity between any two pixels of a feature map. The memory required is beyond the testing desktop can provide. Actually, NLN performs significantly slower than ACNet and CNN according to our observation.  

\begin{figure}[t]
\centering
\includegraphics[width = 1\columnwidth]{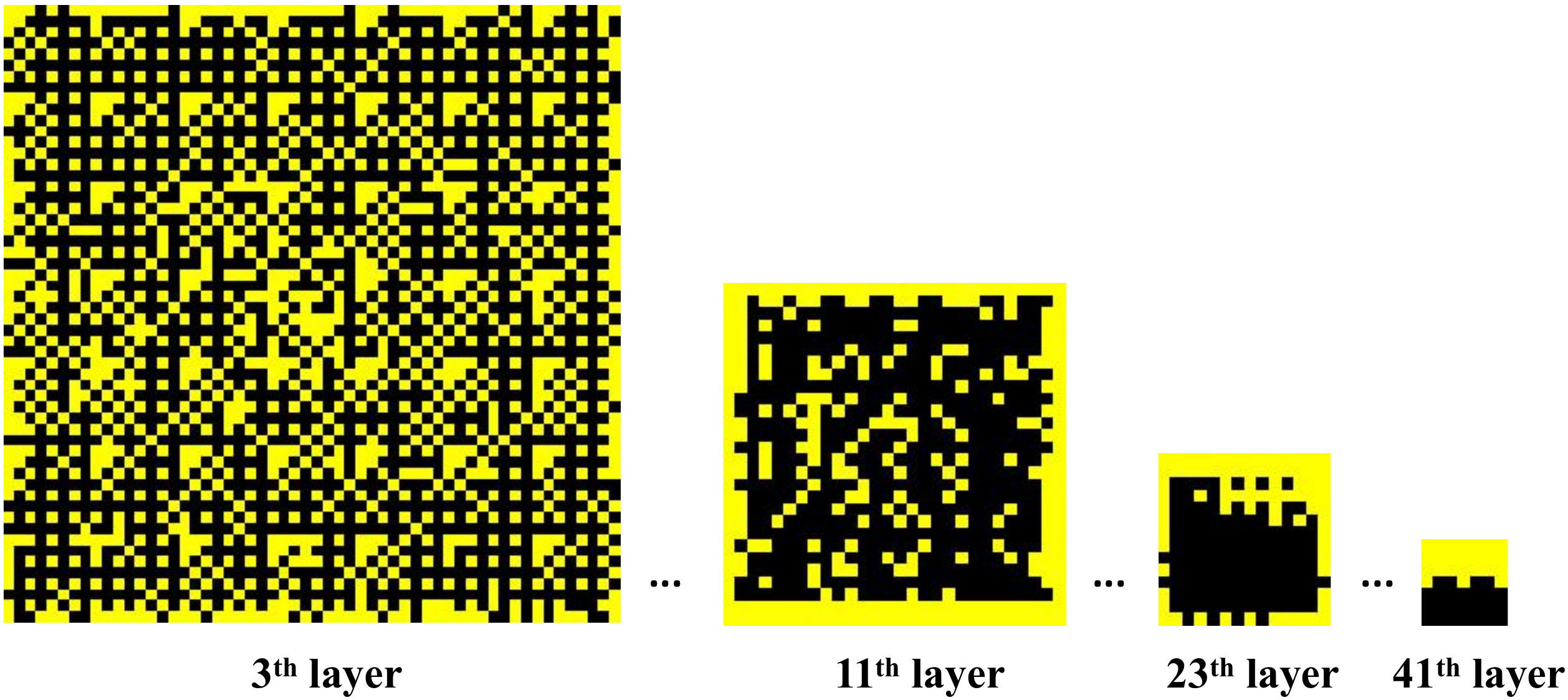}
\caption{\small{Visualization of the nodes with different types of inference generated by our ACNet, which is trained on ImageNet. One node painted by the yellow color indicates its the output of the global inference from the preceding layer (i.e., it connects to all nodes in the preceding layer), while the opposite black nodes indicate the outputs of the local inference from the preceding layer.}}\label{fig:connection}
\vspace{-11pt}
\end{figure}

\textbf{Visualization of importance degrees.} The importance degrees in each ACNet layer are visualized in Fig. \ref{fig:connection}, from which we have two observations. First, the importance degrees differ from pixel to pixel. This is due to the global and local inference are pixel-aware, i.e. different pixels have different inference modes. Second, the importance degrees also differs from layer to layer -- there is much more global inference in lower-level layers than in higher-level layers. {Although CNN somewhat can capture a few global dependencies in high-level layers by stacking a number of local convolutional layer, it} has difficulties in local inference in lower-level layers, as shown in Fig. \ref{fig:connection}. Fortunately, our ACNet provides compensatory global inference for these lower-level layers.
Overall, examining the necessity of global inference in lower layer discloses interesting characteristics and impacts in DNNs, and sheds light on model design in many research fields.

{\subsection{Analysis on CIFAR10}}

As the ImageNet-$1k$ dataset is quite large and the training from scratch is extremely time-consuming, we conduct more ablation studies on CIFAR10 \cite{krizhevsky2009learning} classification benchmark to deeply analyze ACNet.
CIFAR-10 consists of 50k training images and 10k testing image in 10 classes. The presented experiments are trained on the training set and evaluated on the testing set as \cite{he2016deep}. Our focus is to analyze the components of ACNet instead of achieving the state-of-the-art results, therefore we use the representative ResNet32 proposed in \cite{he2016deep}. All the implementation details {and experiment settings are the same as} \cite{he2016deep,wang2018kalman,bkn}.

\textbf{The role of global inference.}
We first evaluate the effectiveness of global inference y constructing two different networks, \ie~with and without the third term $\sum_{\forall j} q_{ij}(\mathbf{x}_j \mathbf{w})$ in Eqn.\ref{eq:gann}.
As shown in Table ~\ref{tab:cora_ablation},
without global inference, ACNet has a performance degradation of 1.1\%.
As we know,  ACNet without global inference equals to CNN. This comparison verifies the superiority of ACNet over CNN.


\begin{table}
\caption{\small{Ablation studies on CIFAR10.}}
\small
\centering
\begin{tabular}{ccc}
\toprule
Method & Error (\%)  \\
\midrule
\textbf{Standard ACNet} & 6.0$^{\downarrow0.0}$ \\
\midrule
w/o global inference & 7.1$^{\downarrow1.1}$ \\
w/o local inference & 24.0$^{\downarrow18}$ \\
Fixed global+local &  6.8$^{\downarrow0.8}$\\
\bottomrule
\end{tabular}\label{tab:cifar_ablation}
\vspace{-11pt}
\end{table}

\textbf{The role of local inference}.
Next, we investigate the necessity of local inference.
 In Table \ref{tab:cora_ablation}, we compare two operations, \ie~ with and without local inference.
Table \ref{tab:cora_ablation} shows that equipped with local inference, ACNet has a significant performance gain of 18\%, verifying the contribution of local inference.
This is natural in the image domain. The lack of local inference leads to neglecting some critical information. Intuitively,
we can easily represent an image as an adjacent matrix. But we can never recover the original image from the adjacent matrix.
demonstrating an information loss by discarding the local inference.

\textbf{Adaptively global+local vs fixed global+local.}
Next, we investigate the necessity of adaptively switching between global and local inference. We fixed the importance degrees $\alpha$, $\beta$ and $\gamma$ as constant variables, forming the \emph{fixed global+local} version of ACNet.
We have an \textbf{interesting} observation in Table \ref{tab:cora_ablation}: imposing global information to every pixel 
has poorer performance than adaptively adding global information (6.0\% \emph{vs} 6.8\%).
In other words, the global inference is unessential for every pixel, because it may hurt the training.
This implies the superiority of adaptively connected neural networks over the fully non-local networks.

\vspace{6pt}
\subsection{COCO Object Detection and Segmentation}

We have demonstrated the adaptive inference capacity of ACNet in ImageNet classification task, whose receptive filed is quite large due to 5 times of subsampling and a global pooling. Next, we investigate an inevitable smaller receptive field task, i.e. COCO 2017 detection \& segmentation task \cite{lin2014microsoft}. These computer vision tasks in general benefit from higher-resolution input and output. Therefore, the global pooling and some subsampling are removed from the backbone of ResNet50, leading to a smaller receptive filed. As a result, the adaptively global and local inference is in desire.

We finetune the models trained on ImageNet \cite{russakovsky2015imagenet} for transferring to detection and segmentation. The batch normalization parameters are frozen during the finetuning.

We conduct experiments on the Mask RCNN baselines \cite{he2017mask} using a ResNet50-FPN backbone. We replace CNN layers with ACNet layers. The models are trained in the COCO train2017 set and evaluated in the COCO val2017 set.
We use the standard training setting following the COCO model zoo. 
We report the standard COCO metrics of Average Precision (AP) for bounding box detection (AP$^{bbox}$) and instance segmentation (AP$^{mask}$).

Table \ref{tab:coco_seg} shows the comparison of ACNet \emph{vs} NLN \emph{vs} CNN. ACNet improves over CNN by 1.5\% box AP and 0.6\% mask AP. This may be contributed to the fact that CNN lacks adaptive inference capacity.
We have also found NLN is 0.5\% box AP worse than ACNet. In summary, although NLN is also suitable global inference, its representational power is slightly weaker than ACNet according to our current evaluation. The inferiority of NLN is attributed to the over-globalization. Specifically, the redundant global context may hurt but NOT help the model learning. This phenomenon has also been observed experimentally by \cite{zhu2018deformable} and theoretically by \cite{lopez2017discovering}, confirming that the over-globalization is a shortcoming of NLN.

\begin{table}
\caption{\small{Detection and segmentation ablation studies on COCO2017 using Mask RCNN.}}
\small
\centering
\begin{tabular}{l|c|c}
\toprule
 backbone & AP$^{bbox}$ & AP$^{mask}$ \\
\midrule
CNN & 38.0$^{\uparrow0.0}$ & 34.6$^{\uparrow0.0}$ \\
NLN & 39.0$^{\uparrow1.0}$  &  35.5$^{\uparrow0.9}$ \\
\textbf{ACNet} & \textbf{39.5}$^{\uparrow1.5}$  & \textbf{35.2}$^{\uparrow0.6}$  \\
\bottomrule
\end{tabular}\label{tab:coco_seg}
\vspace{-11pt}
\end{table}

\subsection{CUHK03 Person Re-identification}
{To demonstrate the good generalization performance of our proposed ACNet on the other recognition tasks, we have conducted the extensive experiments on the person re-identification challenge, which refers to} the problem of re-identifying individuals across cameras. {Though quite challenging, person re-identification is fundamental and beneficial from many applications in video surveillance for keeping the security of safety of the whole society \cite{wang2018p2snet,zhuo2018occluded,wang2018spatial,liang2018m2m,wang2019discovering,ding2015deep,wang2016dari,lin2017cross,li2018distance,li2015deep}.}

\textbf{Dataset.} We conduct experiments on the CUHK03 dataset \cite{li2014deepreid}, which is one of the largest databases for person re-identification. This database contains 14,096 images of 1,467 pedestrians. Each person is observed by two disjoint camera views and is shown in 4.8 images on average in each view. We follow the \textbf{new} standard setting of using CUHK03 \cite{zhong2017random}, where 767 individuals are regarded as the training set and another 700 individuals are considered as the testing set without sharing the same individuals.

\textbf{Evaluation metric.} For the evaluation, the testing set is further divided into a gallery set of images and a probe set. We use the standard rank-1 as the evaluation metric.

\begin{table}[t]
\small
\caption{{Comparison on a Person Re-identification task (CUHK03, where `bs' denotes batch size.)}}
\centering
\begin{tabular}{l|cccc}
\toprule
 & Rank-1  \\
\midrule
BOW+XQDA \cite{zheng2015scalable} & 6.4  \\
PUL \cite{fan2017unsupervised} & 9.1 \\
LOMO+XQDA \cite{liao2015person} & 12.8  \\
IDE \cite{zheng2016person} & 21.3  \\
IDE+DaF \cite{yu2017divide} & 26.4  \\
IDE+XQ.+Re-ranking \cite{zhong2017re} & 34.7 \\
PAN  & 36.3 \\
DPFL \cite{chen2017person} & 40.7  \\
SVDNet \cite{sun2017svdnet} & 41.5  \\
TriNet + Era. \cite{zhong2017random} & 55.5  \\
\hline
TriNet + Era.(Our reproduction) & 62.0$^{\uparrow{0.0}}$  \\
\textbf{TriNet + Era. + ACNet} &  64.3$^{\uparrow{2.3}}$ \\
\hline
{TriNet + Era. + reranking(bs = 32) } &  61.2$^{\uparrow{0.0}}$\\
\textbf{TriNet + Era. + reranking + ACNet(bs = 32)} & 64.8$^{\uparrow{3.6}}$ \\
\bottomrule
\end{tabular}\label{tab:cuhk03}
\vspace{-11pt}
\end{table}

\textbf{Result Analysis.} In Table \ref{tab:cuhk03}, we compare with the current best models. A total of 11 representative state-of-the-art methods, BOW+XQDA \cite{zheng2015scalable}, PUL \cite{fan2017unsupervised}, LOMO+XQDA \cite{liao2015person}, IDE \cite{zheng2016person}, IDE+DaF \cite{yu2017divide}, IDE+XQ.+Re-ranking \cite{zhong2017re}, PAN, DPFL \cite{chen2017person}, and the newly proposed methods SVDNet \cite{sun2017svdnet}, TriNet + Era. \cite{zhong2017random}, and TriNet + Era. + Reranking \cite{zhong2017random} ,  are used as the competing methods. All the settings of the above methods are consistent with the common training settings as \cite{zhong2017random}.
ACNet has achieved a new state-of-the-art performance. Specifically, ACNet achieves a rank-1 accuracy of 64.8\%.
We can also observe that ACNet surpasses its baseline by a clear margin ( 3.6\%, Table \ref{tab:cuhk03}). This verifies the effectiveness of ACNet on the person re-identification task.

\vspace{11pt}
\subsection{Analysis on Cora : a Non-Euclidean Domain}

A common form of graph-structured data is a network of documents. For example, scientific documents in a database are related to each other through citations and references. Administrators of such large networks may desire to automatically label documents according to their relationships to the remainder of the literature. To demonstrate the compatibility of ACNet for non-Euclidean data, we adapt our proposed ACNet to tackle such a vertex classification task on the Cora benchmark~\cite{sen2008collective}, which is a large network of scientific publications connected through citations. The vertex features, in this case, are binary word vectors that indicate the presence of a word from a dictionary of 1,433 unique words. There are 2708 publications classified under 7 different categories - case-based, genetic algorithms, neural networks, probabilistic methods, reinforcement learning, rule learning, and theory. There is an edge connection from a cited article to a citing article and another edge connection from a citing article to a cited article. These edge features are also binary representations. 
We use a quite simple architecture following \cite{kipf2017semi}, which only contains two graph convolutional layers. The first layer is used for feature learning, and the second layer is used for classifier learning. We replace the first graph convolutional layer in \cite{kipf2017semi} with our ACNet layer. Note that in our ACNet $\alpha,\beta,\gamma$ is defined by using Eqn. \ref{eq:unique}. Considering the Cora dataset is quite small-scale, we let $\alpha=0, \beta = 1$ and only learn $\gamma$ to avoid overfitting. We perform 10-fold cross validations to form the training and test set for a fair comparison as the majority of methods~\cite{belkin2006manifold,kipf2017semi} did.

\begin{table}
\caption{\small{Comparison with state-of-the-art on Cora document classification dataset.}}
\small
\centering
\begin{tabular}{c|ccc}
\toprule
{Method} & {Accuracy (\%)} \\
\midrule
{ManiReg} \cite{belkin2006manifold} & {59.5} \\
{SemiEmb} \cite{weston2012deep} & {59.0}  \\
{LP} \cite{zhu2003semi} & {68.0}  \\
{DeepWalk} \cite{perozzi2014deepwalk} &{67.2} \\
{ICA} \cite{lu2003link} & {75.1} \\
{Planetoid*} \cite{yang2016revisiting} & {75.7}  \\
{Graph-CNN} \cite{kipf2017semi} & {81.5} \\
{MoNet} \cite{monti2017geometric} & {81.7} \\
{GAT} \cite{velickovic2017graph} & 83.0 \\
LGCN \cite{gao2018large}& 83.3\\
Dual GCN \cite{zhuang2018dual} & 83.5\\
\hline
{\textbf{ACNet}} & \textbf{83.5} \\
\bottomrule
\end{tabular}\label{tab:cora_sota}
\vspace{-11pt}
\end{table}

\textbf{Comparisons with the state-of-the-art methods.}
We first compare with the current best models. A total of 11 representative state-of-the-art methods, i.e., ManiReg \cite{belkin2006manifold}, SemiEmb \cite{weston2012deep}, LP \cite{zhu2003semi}, DeepWalk \cite{perozzi2014deepwalk}, ICA \cite{lu2003link}, Planetoid \cite{yang2016revisiting}, the newly proposed methods Graph-CNN \cite{kipf2017semi}, MoNet \cite{monti2017geometric}, GAT \cite{velickovic2017graph}, LGCN \cite{gao2018large}, and Dual GCN \cite{zhuang2018dual} are used as competing methods.
Table. \ref {tab:cora_sota} shows that ACNet achieves comparable performance to the best of all competitive methods, e.g., Dual GCN \cite{zhuang2018dual} (83.5\%). This comparison once again verifies the  generalization performance of ACNet. Next, we investigate which component of ACNet contributes to the non-Euclidean data to shed light on future researches.

\textbf{The role of global inference.}
We first evaluate the effectiveness of global inference by constructing two different networks, \ie~with and without the third term $\sum_{\forall j} q_{ij}(\mathbf{x}_j \mathbf{w})$ in Eqn.\ref{eq:gann}.
As shown in Table ~\ref{tab:cora_ablation},
without global inference, ACNet has a performance degradation of 1.4\%. This is reasonable for a document categorization problem like Cora. A document categorization problem is slightly different from a conventional image classification one because each article is not isolated. All the articles are connected with each other in the form of citations. In this sense, a document categorization problem is more like a semantic image segmentation problem in computer vision. Therefore global inference in ACNet is essential for Cora.

\begin{table}
\caption{\small{Ablation studies on Cora document classification dataset.}}
\small
\centering
\begin{tabular}{ccc}
\toprule
Method & Accuracy (\%)  \\
\midrule
\textbf{Standard ACNet} & \textbf{83.5}$^{\downarrow0.0}$ \\
\midrule
w/o global inference & 82.1$^{\downarrow1.4}$\\
w/o local inference & 76.3$^{\downarrow7.2}$ \\
w/o position encoding & 83.0$^{\downarrow0.5}$ \\
Fixed global+local & 82.7$^{\downarrow0.8}$ \\
\bottomrule
\end{tabular}\label{tab:cora_ablation}
\vspace{-11pt}
\end{table}

\textbf{The role of local inference}.
Next, we investigate the necessity of local inference.
 In Table \ref{tab:cora_ablation}, we compare two operations, \ie~ with and without local inference.
Table \ref{tab:cora_ablation} shows that equipped with local inference, ACNet obtains a gain of 7.2\%, verifying the contribution of local inference. The lack of local inference leads to neglecting some critical information. Specifically, each article in Cora cites several other articles, as well as being cited by other articles. Actually, the citing articles and the cited articles may share the same category with the given article. Without local inference, ACNet cannot capture the citation information. {The performance degradation of ``w/o local inference'' may be due to ignoring this knowledge.}

\textbf{Adaptively global+local vs fixed global+local.}
We fixed the importance degrees $\alpha$, $\beta$ and $\gamma$ as constant variable, forming the \emph{fixed global+local} version of ACNet. Similar to the CIFAR10 case, the results in Table \ref{tab:cora_ablation} confirms the effectiveness of adaptively global and local inference, with a gain of 0.8\%. The reason can be attributed to the property of the document. Some article can be easier to categorize when considered in local range than in wide range. For example, at first, we can easily categorize the reinforcement-learning-based article into the ``reinforcement learning'' area. But after reading more and more article, we may confuse it with ``neural networks'' area with the emergence of deep reinforcement learning.

\textbf{The role of position encoding.}
At last, we investigate the impact of position encoding.
We remove the position encoding in Eqn.\ref{eq:gann} to obtain the counterpart. Table \ref{tab:cora_ablation} shows that without the position encoding, ACNet suffers a performance drop of 0.5\%. This is because the non-Euclidean data is unstructured compared with the Euclidean data. Without a position encoding, the non-Euclidean data is with too many degrees of freedom (i.e., the same graph data may have different representations because theoretically, a graph has endless isomorphic graphs). This freedom leads to lower learning efficiency. By introducing the position encoding the training inefficiency has been alleviated


\section{Conclusion}

This paper presented a concise ACNet to be a promising substitute for overcoming the limitations of widely used deep CNNs without losing their strengths in feature learning. Specifically, ACNet advances in adaptively switching between global and local inference in a flexible and pure data-driven manner. We further applied our proposed ACNet for the recognition tasks of both Euclidean data and non-Euclidean data. Extensive experimental analyses from a variety of aspects justify the superiority of ACNet. In the future, we will extend our work to be suitable for more general tasks to demonstrate its superiority.

\section*{Acknowledgement}
\noindent
This work was supported in part by the National Key Research and Development Program of China under Grant No. 2018YFC0830103 and 2016YFB1001004, in part by National High Level Talents Special Support Plan (Ten Thousand Talents Program), and in part by National Natural Science Foundation of China (NSFC) under Grant No. 61622214,  61836012, and 61876224.

{\small
\bibliographystyle{ieee}
\bibliography{acnet_arxiv}

\begin{thebibliography}{10}\itemsep=-1pt

\bibitem{belkin2006manifold}
Mikhail Belkin, Partha Niyogi, and Vikas Sindhwani.
\newblock Manifold regularization: A geometric framework for learning from
  labeled and unlabeled examples.
\newblock {\em Journal of machine learning research}, 7(Nov):2399--2434, 2006.

\bibitem{chen2017rethinking}
Liang-Chieh Chen, George Papandreou, Florian Schroff, and Hartwig Adam.
\newblock Rethinking atrous convolution for semantic image segmentation.
\newblock {\em arXiv preprint arXiv:1706.05587}, 2017.

\bibitem{chen2018graph}
Yunpeng Chen, Marcus Rohrbach, Zhicheng Yan, Shuicheng Yan, Jiashi Feng, and
  Yannis Kalantidis.
\newblock Graph-based global reasoning networks.
\newblock {\em arXiv preprint arXiv:1811.12814}, 2018.

\bibitem{chen2017person}
Yanbei Chen, Xiatian Zhu, and Shaogang Gong.
\newblock Person re-identification by deep learning multi-scale
  representations.
\newblock In {\em CVPR}, pages 2590--2600, 2017.

\bibitem{deformableCNN17ICCV}
Jifeng Dai, Haozhi Qi, Yuwen Xiong, Yi Li, Guodong Zhang, Han Hu, and Yichen
  Wei.
\newblock Deformable convolutional networks.
\newblock In {\em ICCV}, 2017.

\bibitem{ding2015deep}
Shengyong Ding, Liang Lin, Guangrun Wang, and Hongyang Chao.
\newblock Deep feature learning with relative distance comparison for person
  re-identification.
\newblock {\em Pattern Recognition}, 48(10):2993--3003, 2015.

\bibitem{fan2017unsupervised}
Hehe Fan, Liang Zheng, and Yi Yang.
\newblock Unsupervised person re-identification: Clustering and fine-tuning.
\newblock {\em arXiv preprint arXiv:1705.10444}, 2017.

\bibitem{gao2018large}
Hongyang Gao, Zhengyang Wang, and Shuiwang Ji.
\newblock Large-scale learnable graph convolutional networks.
\newblock In {\em Proceedings of the 24th ACM SIGKDD International Conference
  on Knowledge Discovery \& Data Mining}, pages 1416--1424. ACM, 2018.

\bibitem{he2017mask}
Kaiming He, Georgia Gkioxari, Piotr Doll{\'a}r, and Ross Girshick.
\newblock Mask r-cnn.
\newblock In {\em Computer Vision (ICCV), 2017 IEEE International Conference
  on}, pages 2980--2988. IEEE, 2017.

\bibitem{resnet}
Kaiming He, Xiangyu Zhang, Shaoqing Ren, and Jian Sun.
\newblock Deep residual learning for image recognition.
\newblock In {\em CVPR}, 2016.

\bibitem{he2016deep}
Kaiming He, Xiangyu Zhang, Shaoqing Ren, and Jian Sun.
\newblock Deep residual learning for image recognition.
\newblock In {\em Proceedings of the IEEE conference on computer vision and
  pattern recognition}, pages 770--778, 2016.

\bibitem{matrixCapsule}
Geoffrey~E Hinton, Sara Sabour, and Nicholas Frosst.
\newblock Matrix capsules with em routing.
\newblock In {\em ICLR}, 2018.

\bibitem{huang2017densely}
Gao Huang, Zhuang Liu, Kilian~Q Weinberger, and Laurens van~der Maaten.
\newblock Densely connected convolutional networks.
\newblock In {\em CVPR}, volume~1, page~3, 2017.

\bibitem{ioffe2015batch}
Sergey Ioffe and Christian Szegedy.
\newblock Batch normalization: Accelerating deep network training by reducing
  internal covariate shift.
\newblock In {\em International conference on machine learning}, pages
  448--456, 2015.

\bibitem{kipf2017semi}
Thomas~N Kipf and Max Welling.
\newblock Semi-supervised classification with graph convolutional networks.
\newblock {\em ICLR}, 2017.

\bibitem{krizhevsky2009learning}
Alex Krizhevsky and Geoffrey Hinton.
\newblock Learning multiple layers of features from tiny images.
\newblock 2009.

\bibitem{BP2}
Yann Lecun.
\newblock {\em PhD thesis: Modeles connexionnistes de l'apprentissage
  (connectionist learning models)}.
\newblock Universite P. et M. Curie (Paris 6), 6 1987.

\bibitem{lecun1990handwritten}
Yann LeCun, Bernhard~E Boser, John~S Denker, Donnie Henderson, Richard~E
  Howard, Wayne~E Hubbard, and Lawrence~D Jackel.
\newblock Handwritten digit recognition with a back-propagation network.
\newblock In {\em Advances in neural information processing systems}, pages
  396--404, 1990.

\bibitem{lenet}
Y. LeCun, L. Bottou, Y. Bengio, and P. Haffner.
\newblock Gradient-based learning applied to document recognition.
\newblock {\em Proceedings of the IEEE}, 11(6):2278–2324, 1998.

\bibitem{li2014deepreid}
Wei Li, Rui Zhao, Tong Xiao, and Xiaogang Wang.
\newblock Deepreid: Deep filter pairing neural network for person
  re-identification.
\newblock In {\em Proceedings of the IEEE Conference on Computer Vision and
  Pattern Recognition}, pages 152--159, 2014.

\bibitem{li2017aognets}
Xilai Li, Tianfu Wu, Xi Song, and Hamid Krim.
\newblock Aognets: Deep and-or grammar networks for visual recognition.
\newblock {\em arXiv preprint arXiv:1711.05847}, 2017.

\bibitem{li2015deep}
Ya Li, Guangrun Wang, Liang Lin, and Huiyou Chang.
\newblock A deep joint learning approach for age invariant face verification.
\newblock In {\em CCF Chinese Conference on Computer Vision}, pages 296--305.
  Springer, 2015.

\bibitem{li2018distance}
Ya Li, Guangrun Wang, Lin Nie, Qing Wang, and Wenwei Tan.
\newblock Distance metric optimization driven convolutional neural network for
  age invariant face recognition.
\newblock {\em Pattern Recognition}, 75:51--62, 2018.

\bibitem{liang2018m2m}
Wenqi Liang, Guangcong Wang, Jianhuang Lai, and Junyong Zhu.
\newblock M2m-gan: Many-to-many generative adversarial transfer learning for
  person re-identification.
\newblock {\em arXiv preprint arXiv:1811.03768}, 2018.

\bibitem{liao2015person}
Shengcai Liao, Yang Hu, Xiangyu Zhu, and Stan~Z Li.
\newblock Person re-identification by local maximal occurrence representation
  and metric learning.
\newblock In {\em CVPR}, pages 2197--2206, 2015.

\bibitem{lin2017cross}
Liang Lin, Guangrun Wang, Wangmeng Zuo, Xiangchu Feng, and Lei Zhang.
\newblock Cross-domain visual matching via generalized similarity measure and
  feature learning.
\newblock {\em IEEE transactions on pattern analysis and machine intelligence},
  39(6):1089--1102, 2017.

\bibitem{lin2014microsoft}
Tsung-Yi Lin, Michael Maire, Serge Belongie, James Hays, Pietro Perona, Deva
  Ramanan, Piotr Doll{\'a}r, and C~Lawrence Zitnick.
\newblock Microsoft coco: Common objects in context.
\newblock In {\em European conference on computer vision}, pages 740--755.
  Springer, 2014.

\bibitem{lopez2017discovering}
David Lopez-Paz, Robert Nishihara, Soumith Chintala, Bernhard Scholkopf, and
  L{\'e}on Bottou.
\newblock Discovering causal signals in images.
\newblock In {\em Proceedings of the IEEE Conference on Computer Vision and
  Pattern Recognition}, pages 6979--6987, 2017.

\bibitem{lu2003link}
Qing Lu and Lise Getoor.
\newblock Link-based classification.
\newblock In {\em Proceedings of the 20th International Conference on Machine
  Learning (ICML-03)}, pages 496--503, 2003.

\bibitem{monti2017geometric}
Federico Monti, Davide Boscaini, Jonathan Masci, Emanuele Rodola, Jan Svoboda,
  and Michael~M Bronstein.
\newblock Geometric deep learning on graphs and manifolds using mixture model
  cnns.
\newblock In {\em Proc. CVPR}, volume~1, page~3, 2017.

\bibitem{peng2017large}
Chao Peng, Xiangyu Zhang, Gang Yu, Guiming Luo, and Jian Sun.
\newblock Large kernel matters—improve semantic segmentation by global
  convolutional network.
\newblock In {\em Computer Vision and Pattern Recognition (CVPR), 2017 IEEE
  Conference on}, pages 1743--1751. IEEE, 2017.

\bibitem{perozzi2014deepwalk}
Bryan Perozzi, Rami Al-Rfou, and Steven Skiena.
\newblock Deepwalk: Online learning of social representations.
\newblock In {\em Proceedings of the 20th ACM SIGKDD international conference
  on Knowledge discovery and data mining}, pages 701--710. ACM, 2014.

\bibitem{ronneberger2015u}
Olaf Ronneberger, Philipp Fischer, and Thomas Brox.
\newblock U-net: Convolutional networks for biomedical image segmentation.
\newblock In {\em International Conference on Medical image computing and
  computer-assisted intervention}, pages 234--241. Springer, 2015.

\bibitem{BP}
D. Rumelhart, G. Hinton, and R. Williams.
\newblock Learning internal representations by backpropagating errors.
\newblock {\em Parallel distributed processing: Explorations in the
  microstructure of cognition}, 1986.

\bibitem{russakovsky2015imagenet}
Olga Russakovsky, Jia Deng, Hao Su, Jonathan Krause, Sanjeev Satheesh, Sean Ma,
  Zhiheng Huang, Andrej Karpathy, Aditya Khosla, Michael Bernstein, et~al.
\newblock Imagenet large scale visual recognition challenge.
\newblock {\em International Journal of Computer Vision}, 115(3):211--252,
  2015.

\bibitem{capsule}
Sara Sabour, Nicholas Frosst, and Geoffrey Hinton.
\newblock Dynamic routing between capsules.
\newblock In {\em NIPS}, 2017.

\bibitem{sen2008collective}
Prithviraj Sen, Galileo Namata, Mustafa Bilgic, Lise Getoor, Brian Galligher,
  and Tina Eliassi-Rad.
\newblock Collective classification in network data.
\newblock {\em AI magazine}, 29(3):93, 2008.

\bibitem{graphcnn}
Felipe~Petroski Such, Shagan Sah, Miguel Dom{\'{\i}}nguez, Suhas Pillai, Chao
  Zhang, Andrew Michael, Nathan~D. Cahill, and Raymond~W. Ptucha.
\newblock Robust spatial filtering with graph convolutional neural networks.
\newblock {\em J. Sel. Topics Signal Processing}, 11(6):884--896, 2017.

\bibitem{sun2017svdnet}
Yifan Sun, Liang Zheng, Weijian Deng, and Shengjin Wang.
\newblock Svdnet for pedestrian retrieval.
\newblock {\em arXiv preprint arXiv:1703.05693}, 2017.

\bibitem{selfattention17NIPS}
A. Vaswani, N. Shazeer, N. Parmar, J. Uszkoreit, L. Jones, A.~N. Gomez, L.
  Kaiser, and I. Polosukhin.
\newblock Attention is all you need.
\newblock In {\em NIPS}, 2017.

\bibitem{velickovic2017graph}
Petar Velickovic, Guillem Cucurull, Arantxa Casanova, Adriana Romero, Pietro
  Lio, and Yoshua Bengio.
\newblock Graph attention networks.
\newblock {\em arXiv preprint arXiv:1710.10903}, 1(2), 2017.

\bibitem{wang2018spatial}
Guangcong Wang, Jianhuang Lai, Peigen Huang, and Xiaohua Xie.
\newblock Spatial-temporal person re-identification.
\newblock {\em arXiv preprint arXiv:1812.03282}, 2018.

\bibitem{wang2018p2snet}
Guangcong Wang, Jianhuang Lai, and Xiaohua Xie.
\newblock P2snet: Can an image match a video for person re-identification in an
  end-to-end way?
\newblock {\em IEEE Transactions on Circuits and Systems for Video Technology},
  28(10):2777--2787, 2018.

\bibitem{wang2019discovering}
Guangcong Wang, Jianhuang Lai, Zhenyu Xie, and Xiaohua Xie.
\newblock Discovering underlying person structure pattern with relative local
  distance for person re-identification.
\newblock {\em arXiv preprint arXiv:1901.10100}, 2019.

\bibitem{wang2016dari}
Guangrun Wang, Liang Lin, Shengyong Ding, Ya Li, and Qing Wang.
\newblock Dari: Distance metric and representation integration for person
  verification.
\newblock In {\em Thirtieth AAAI Conference on Artificial Intelligence}, 2016.

\bibitem{wang2018kalman}
Guangrun Wang, Ping Luo, Xinjiang Wang, Liang Lin, et~al.
\newblock Kalman normalization: Normalizing internal representations across
  network layers.
\newblock In {\em Advances in Neural Information Processing Systems}, pages
  21--31, 2018.

\bibitem{bkn}
Guangrun Wang, Jiefeng Peng, Ping Luo, Xinjiang Wang, and Liang Lin.
\newblock Batch kalman normalization: Towards training deep neural networks
  with micro-batches.
\newblock {\em arXiv preprint arXiv:1802.03133}, 2018.

\bibitem{nonlocalnn17}
Xiaolong Wang, Ross Girshick, Abhinav Gupta, and Kaiming He.
\newblock Non-local neural networks.
\newblock In {\em arXiv:1711.07971 [cs.LG]}, 2017.

\bibitem{weston2012deep}
Jason Weston, Fr{\'e}d{\'e}ric Ratle, Hossein Mobahi, and Ronan Collobert.
\newblock Deep learning via semi-supervised embedding.
\newblock In {\em Neural Networks: Tricks of the Trade}, pages 639--655.
  Springer, 2012.

\bibitem{yang2016revisiting}
Zhilin Yang, William~W Cohen, and Ruslan Salakhutdinov.
\newblock Revisiting semi-supervised learning with graph embeddings.
\newblock {\em arXiv preprint arXiv:1603.08861}, 2016.

\bibitem{yu2017divide}
Rui Yu, Zhichao Zhou, Song Bai, and Xiang Bai.
\newblock Divide and fuse: A re-ranking approach for person re-identification.
\newblock {\em arXiv preprint arXiv:1708.04169}, 2017.

\bibitem{zhao2018psanet}
Hengshuang Zhao, Yi Zhang, Shu Liu, Jianping Shi, Chen Change~Loy, Dahua Lin,
  and Jiaya Jia.
\newblock Psanet: Point-wise spatial attention network for scene parsing.
\newblock In {\em Proceedings of the European Conference on Computer Vision
  (ECCV)}, pages 267--283, 2018.

\bibitem{zheng2015scalable}
Liang Zheng, Liyue Shen, Lu Tian, Shengjin Wang, Jingdong Wang, and Qi Tian.
\newblock Scalable person re-identification: A benchmark.
\newblock In {\em ICCV}, 2015.

\bibitem{zheng2016person}
Liang Zheng, Yi Yang, and Alexander~G Hauptmann.
\newblock Person re-identification: Past, present and future.
\newblock {\em arXiv preprint arXiv:1610.02984}, 2016.

\bibitem{zhong2017re}
Zhun Zhong, Liang Zheng, Donglin Cao, and Shaozi Li.
\newblock Re-ranking person re-identification with k-reciprocal encoding.
\newblock {\em arXiv preprint arXiv:1701.08398}, 2017.

\bibitem{zhong2017random}
Zhun Zhong, Liang Zheng, Guoliang Kang, Shaozi Li, and Yi Yang.
\newblock Random erasing data augmentation.
\newblock {\em arXiv preprint arXiv:1708.04896}, 2017.

\bibitem{zhu2003semi}
Xiaojin Zhu, Zoubin Ghahramani, and John~D Lafferty.
\newblock Semi-supervised learning using gaussian fields and harmonic
  functions.
\newblock In {\em Proceedings of the 20th International conference on Machine
  learning (ICML-03)}, pages 912--919, 2003.

\bibitem{zhu2018deformable}
Xizhou Zhu, Han Hu, Stephen Lin, and Jifeng Dai.
\newblock Deformable convnets v2: More deformable, better results.
\newblock {\em arXiv preprint arXiv:1811.11168}, 2018.

\bibitem{zhuang2018dual}
Chenyi Zhuang and Qiang Ma.
\newblock Dual graph convolutional networks for graph-based semi-supervised
  classification.
\newblock In {\em Proceedings of the 2018 World Wide Web Conference on World
  Wide Web}, pages 499--508. International World Wide Web Conferences Steering
  Committee, 2018.

\bibitem{zhuo2018occluded}
Jiaxuan Zhuo, Zeyu Chen, Jianhuang Lai, and Guangcong Wang.
\newblock Occluded person re-identification.
\newblock In {\em 2018 IEEE International Conference on Multimedia and Expo
  (ICME)}, pages 1--6. IEEE, 2018.

\end{thebibliography}
}

\end{document}